\documentclass[runningheads]{llncs}
\usepackage{graphicx}
\usepackage{tikz}
\usepackage{comment}
\usepackage{amsmath,amssymb} % define this before the line numbering.
\usepackage{color}
\usepackage{subfigure}
\usepackage{bm}
\usepackage{bbding}
\usepackage{arydshln}
\usepackage{setspace}
\usepackage{multirow}
\usepackage{float}
\usepackage{wrapfig}

\usepackage[accsupp]{axessibility}  % Improves PDF readability for those with disabilities.

\usepackage[width=122mm,left=12mm,paperwidth=146mm,height=193mm,top=12mm,paperheight=217mm]{geometry}

\begin{document}
% \renewcommand\thelinenumber{\color[rgb]{0.2,0.5,0.8}\normalfont\sffamily\scriptsize\arabic{linenumber}\color[rgb]{0,0,0}}
% \renewcommand\makeLineNumber {\hss\thelinenumber\ \hspace{6mm} \rlap{\hskip\textwidth\ \hspace{6.5mm}\thelinenumber}}
% \linenumbers
\pagestyle{headings}
\mainmatter
\def\ECCVSubNumber{****}  % Insert your submission number here

\title{Controllable Augmentations for Video Representation Learning} % Replace with your title

% INITIAL SUBMISSION 
\begin{comment}
\titlerunning{ECCV-22 submission ID \ECCVSubNumber} 
\authorrunning{ECCV-22 submission ID \ECCVSubNumber} 
\author{Anonymous ECCV submission}
\institute{Paper ID \ECCVSubNumber}
\end{comment}
%******************

% CAMERA READY SUBMISSION
%\begin{comment}
\titlerunning{Controllable Augmentations}
% If the paper title is too long for the running head, you can set
% an abbreviated paper title here
%
\author{Rui Qian\inst{1,2} \and
Weiyao Lin\inst{1} \and
John See\inst{3} \and Dian Li\inst{4}}
\authorrunning{R. Qian et al.}
% First names are abbreviated in the running head.
% If there are more than two authors, 'et al.' is used.
%
\institute{Shanghai Jiao Tong University, China \and The Chinese University of Hong Kong, China \and
Heriot-Watt University, Malaysia \and Tencent, China}
%\end{comment}
%******************
\maketitle

\begin{abstract}
    This paper focuses on self-supervised video representation learning. Most existing approaches follow the contrastive learning pipeline to construct positive and negative pairs by sampling different clips. However, this formulation tends to bias to static background and have difficulty establishing global temporal structures. The major reason is that the positive pairs, i.e., different clips sampled from the same video, have limited temporal receptive field, and usually share similar background but differ in motions. To address these problems, we propose a framework to jointly utilize local clips and global videos to learn from detailed region-level correspondence as well as general long-term temporal relations. Based on a set of controllable augmentations, we achieve accurate appearance and motion pattern alignment through soft spatio-temporal region contrast. Our formulation is able to avoid the low-level redundancy shortcut by mutual information minimization to improve the generalization. We also introduce local-global temporal order dependency to further bridge the gap between clip-level and video-level representations for robust temporal modeling. Extensive experiments demonstrate that our framework is superior on three video benchmarks in action recognition and video retrieval, capturing more accurate temporal dynamics.
\end{abstract}

\section{Introduction}
\label{intro}
Video representation learning is fundamental to video understanding applications, \emph{e.g.}, action recognition~\cite{carreira2017quo,xie2018rethinking}, spatio-temporal detection~\cite{gu2018ava,caba2015activitynet}, video retrieval~\cite{liu2019use,miech2019howto100m}, etc. Traditional supervised learning schemes require large-scale human labeling, and the performance is usually restricted by the granularity of annotations. More precisely, coarse-grained video-level annotations could lead the model to attend to the background~\cite{soomro2012ucf101,carreira2017quo}, while fine-grained annotations greatly facilitate general video analysis but are much more expensive~\cite{goyal2017something,li2018resound}. To solve this problem, unsupervised video representation learning begins to attract more attention. Some early works design diverse pretext tasks to learn the video characteristics in a self-supervised manner~\cite{benaim2020speednet,misra2016shuffle,kim2019self,jenni2020video,xu2019self,wang2020statistic}. Recently, the formulation of contrastive learning further improves the performance by a large margin~\cite{gordon2020watching,qian2020spatiotemporal,wang2020self,yao2020seco}.

\begin{wrapfigure}{r}{0.5\textwidth}
    \centering
    \includegraphics[width=\linewidth]{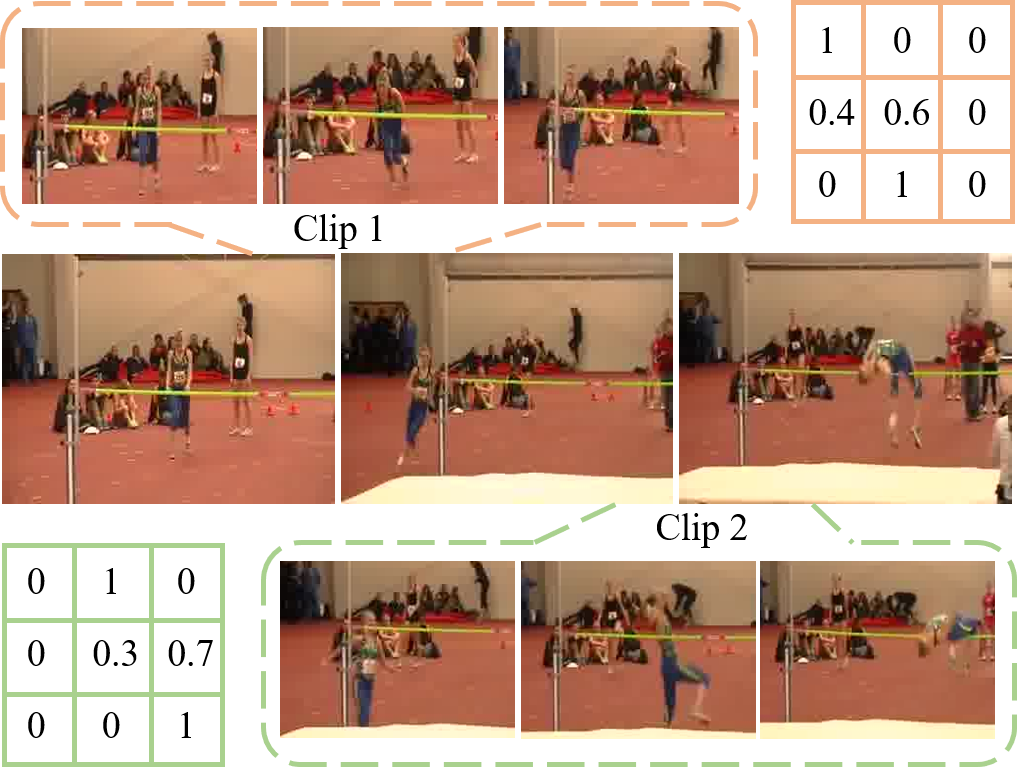}
    \caption{An illustration of clip sampling and temporal correspondence. We show a high-jump video with two sampled clips. Two clips have the same background but different motions: one running and the other jumping. We provide an example of temporal correspondence between clip and video, where we coarsely divide the video (clip) into three segments, the value in the matrix indicates the intersection ratio. The spatial correspondence can be calculated similarly.}
    \vspace{-4mm}
    \label{fig:teaser}
\end{wrapfigure}

A prevalent way for contrastive video representation learning is to sample several clips and regard those from the video as positive pairs~\cite{qian2020spatiotemporal,han2020self,kuang2021video,pan2021videomoco}. However, this formulation has two drawbacks. On one hand, these methods tend to be bias towards static background~\cite{wang2021removing,wang2020enhancing}. This is because the sampled clips mostly share the same background, but there probably exists subtle differences in motions. For example, in Fig.~\ref{fig:teaser}, the video contains a high jump scene. The clip sampled at an early timestamp shows the running action, but that same clip sampled at a later timestamp presents the jumping action. Thus, pulling these two clips closer in the feature space will lead the model neglecting their distinct motions and only attend to the background of the stadium. On the other hand, there remains an obvious gap between clip-level features and video-level representation. The sampled clips have limited temporal receptive field, and thus, cannot provide comprehensive information. For example, Clip 1 in Fig.~\ref{fig:teaser} only shows the momentary process of running. When we jointly leverage the correctly ordered two clips, \emph{i.e.}, the running action happens before jumping, we would be able to understand the original video. Motivated by these observations, we intend to address these problems from two aspects, one is detailed region-level correspondence, the other is general long-term temporal perception.

In this paper, we propose a framework to learn comprehensive appearance and motion patterns in videos. Concretely, we develop a set of controllable augmentations to achieve this goal. First, we use constrained spatio-temporal cropping to sample several local clips from each video such that the clips cover diverse timestamps of the video. Then, based on the cropping parameters, we generate dense spatio-temporal position-wise correspondences between the local clip and global video feature maps. In Fig.~\ref{fig:teaser}, we show a toy example on temporal correspondence, whereby the spatial correspondence is established by employing these soft codes to align features in corresponding regions. In this way, we can match the exact same appearance and motion content, while avoid aligning inconsistent motions between various timestamps. However, there also exist ``shortcuts'' that govern the overlapping regions between local clips and global videos, \emph{e.g.}, the low-level color statistics; these shortcuts could prevent the model from learning useful semantics. To avoid them, we define different intensity levels of color jitter and Gaussian blur augmentations, and regard the samples generated by the same level augmentation as sharing similar low-level attributes. We then minimize the mutual information between them to mitigate the impact of low-level shortcuts on the extracted representation.

To further bridge the gap between clip-level and video-level representations, we intuitively introduce a learning objective to model temporal order dependency between local clips and global video. Particularly, we have access to the temporal order of the sampled clips in accordance to the cropping parameters. With that, we aim to maximize the mutual information between correctly ordered clip features and the global video feature. Through this operation, we facilitate the model's temporal awareness in the pretraining stage.

In summary, our contributions are as follows:
\begin{itemize}
    \item We propose a unified framework to learn video representations from detailed local contrast and general long-term temporal modeling.
    \item We develop controllable augmentations to match the visual contents in corresponding spatio-temporal positions for detailed content alignment, and perform mutual information minimization to avoid low-level shortcut.
    \item We introduce the temporal order dependency between the local clips and global video to enhance general temporal structure modeling.
    \item We achieve superior results on downstream action recognition and video retrieval tasks, while capturing more accurate motion patterns.
\end{itemize}

\section{Related Work}

\subsubsection{Contrastive Learning.}

Recently, contrastive learning has revolutionized self-supervised learning~\cite{he2020momentum,chen2020simple,oord2018representation}. Its core idea is to discriminate different instances by attracting the positive pairs and repelling the negative pairs in feature space~\cite{hadsell2006dimensionality,gutmann2010noise}. Following this, \cite{wu2018unsupervised} formulates the instance discrimination as a non-parametric classification problem. \cite{oord2018representation} proposes to estimate mutual information with InfoNCE loss~\cite{gutmann2010noise}, which leads to easy optimization and fast convergence. Inspired by this, a line of works~\cite{tian2019contrastive,chen2020simple,he2020momentum,hjelm2018learning} adopt this learning objective for image representation learning and show significant improvement on downstream tasks. Later, \cite{xie2021propagate,wang2021dense} develop dense contrastive learning, which performs pixel-level contrast. Compared to instance-level discrimination, dense contrastive learning preserves richer characteristics, and performs better on dense prediction tasks and visual correspondence learning. In our work, we focus on video representation learning. Considering that there exists natural spatio-temporal correspondences in video domain, we propose to utilize it as a self-supervisory signal for spatio-temporal region contrast to learn more comprehensive video representations.

\subsubsection{Video Representation Learning.}

Unlike images, videos contain internal temporal structures that are crucial for video content analysis. To this end, many works~\cite{misra2016shuffle,lee2017unsupervised,xu2019self} have designed various pretext tasks to leverage the natural spatio-temporal correspondence as self-supervisory signals. Some typical pretext tasks include temporal ordering~\cite{misra2016shuffle,xu2019self,yao2020seco}, spatio-temporal puzzles~\cite{kim2019self,wang2020statistic}, colorization~\cite{vondrick2018tracking}, playback speed prediction~\cite{jenni2020video,benaim2020speednet}, temporal cycle-consistency~\cite{wang2019learning,jabri2020space,li2019joint}, and future prediction~\cite{vondrick2016anticipating,villegas2017decomposing,luo2017unsupervised,behrmann2021unsupervised}. There are also some works using cross-modal correspondence for self-supervised pretraining~\cite{alwassel2019self,piergiovanni2020evolving}.
Inspired by the success of contrastive learning in image domain, a series of works extend this pipeline to video domain~\cite{gordon2020watching,qian2020spatiotemporal,wang2020self,liu2021temporal}. Particularly, \cite{han2019video,han2020memory} employ InfoNCE loss for dense future prediction, while~\cite{wang2020self,yang2020video} sample clips of different rates as positive pairs for visual content learning. However, video contrastive learning could lead the model to lay more emphasis on the static scene and focus less on motion~\cite{wang2021removing}. To solve this problem, ~\cite{chen2021rspnet,jenni2020video} propose to integrate contrastive learning with temporal pretext tasks to enhance the temporal awareness. \cite{han2020self,li2021motion} use optical flow to assist motion modeling. In our work, we do not resort to optical flow to enhance motion learning and temporal modeling. Instead, we hypothesize that the underlying reason for static scene bias lies in the positive pair formulation. That is, most existing works use either different frames~\cite{yao2020seco,gordon2020watching} or different clips~\cite{qian2020spatiotemporal,kuang2021video} from the same video as the positive pair, which usually have similar background but possess different motions. Hence, we propose to consider the corresponding regions within local and global views to form accurate positive, concurrently with low-level shortcut elimination, which captures the desired static and dynamic characteristics. In addition, we develop a temporal dependency between these views to bridge the gap between local clip and global video representations, while learning robust temporal structures.

\subsubsection{Local-global Views for Video Representation.}

There have been some works also using local and global views for self-supervised video representation learning~\cite{ma2021contrastive,recasens2021broaden,dave2021tclr,behrmann2021long,kuang2021video}. The major difference between our work and those works lies in the concept of local global views and its target. In our work, ``local global'' means short and long video clips, and the major target is to construct \textit{spatio-temporal overlaps} and formulate a \textit{soft learning objective}, which guides detailed region-level video content alignment. In~\cite{ma2021contrastive}, local global means local fine-grained and global coarse-grained features, which is designed for general audio-visual correspondence. \cite{recasens2021broaden} aims to extrapolate the neighboring video content in global view based on the observation from the local view. TCLR~\cite{tclr} designs a loss function to learn temporal correspondence between local and global clips but still with \textit{hard positive assignment}. \cite{lsfd} employs local global views to decompose stationary and non-stationary features and~\cite{vclr} uses them for segment-based positive sampling.

\section{Method}
\begin{figure*}
    \centering
    \includegraphics[width=\linewidth]{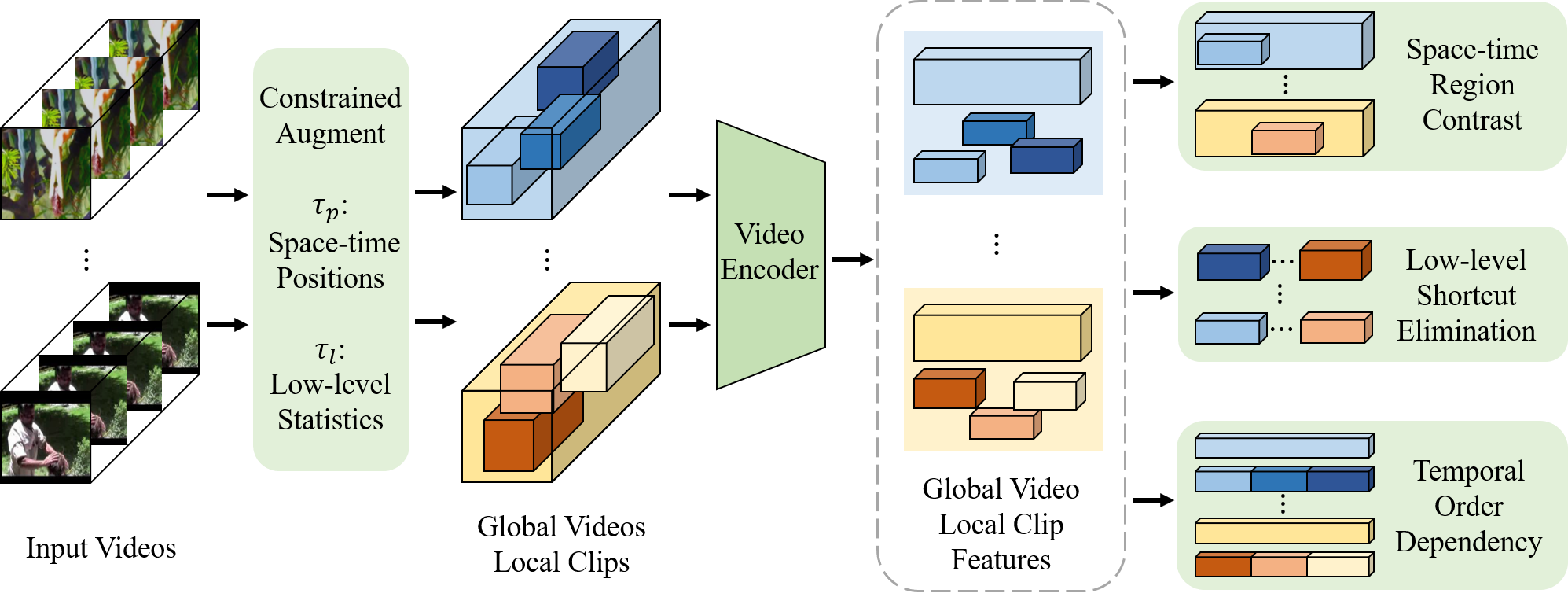}
    \caption{An overview of the proposed local-global composition framework. We define a set of controllable augmentations to generate the global video and local clip input. Based on the extracted features, we perform spatio-temporal region contrative learning for accurate visual content alignment, and minimize the mutual information between samples that low-level statistics to eliminate the shortcut. And we construct the local-global temporal order dependency to bridge the gap between clip-level and video-level features. Note that in this figure, we use cubes to present videos or clips, similar cube color means they derive from the same video, and similar color brightness means two cubes share similar low-level statistics.}
    \label{fig:framework}
\end{figure*}
The core idea of our proposed framework is to enhance self-supervised video representation learning by comprehensive appearance and motion content modeling. As shown in Fig.~\ref{fig:framework}, we utilize a set of controllable augmentations to achieve (1) detailed spatio-temporal region contrast, (2) low-level shortcut elimination and (3) general temporal dependency modeling.

Specifically, we divide the augmentations into two parts: one is spatio-temporal position transformations $\tau_p$ (including crop and horizontal flip), the other is low-level statistic transformations $\tau_l$ (including color jitter and Gaussian blur). Following the data preprocessing pipeline, given a video $v$, we first use $\tau_p$ to sample several local clips and then perform $\tau_l$ to generate the input to the encoder.

\subsection{Spatio-temporal Region Contrast}

Given a video $v$ with temporal length $T$, we first use spatio-temporal cropping to sample $K$ clips \emph{i.e.} $v_k\in\{v_1,v_2,...,v_K\}$, to provide the local feature descriptions. In order to let the sampled clips contain as much information as the original video, we manually constrain the temporal cropping parameters in $\tau_p^k$ to control the central timestamp of $v_k$ in range of $\left[\frac{(k-1)T}{K},\frac{kT}{K}\right]$. In this way, sampled clips cover different temporal segments and they jointly present the rich information in $v$. As mentioned in Sec.~\ref{intro}, there could be inconsistencies in motions between different local clips such that it is not optimal to align the representations between different clips. Hence, we need to figure out the exact corresponding content for feature alignment. To this end, considering that there is natural correspondence between local clips and global video, we leverage $v$ and $v_k$ as two views for feature matching.

For local clip feature extraction, we denote the feature extractor as $f(\cdot)$, and the local clip feature map as $f(v_k)\in\mathbb{R}^{CT_cHW}$, where $C,H,W$ denotes the channel, width and height respectively, $T_c$ denotes the temporal dimension of clip feature map. For global video feature extraction, we perform sparse sampling to represent $v$, and set some convolution layers' temporal stride to 1 to make $f'(v)\in\mathbb{R}^{CT_vHW}$ possess higher temporal resolution, \emph{i.e.}, temporal dimension $T_v>T_c$. Note that $f$ and $f'$ share the same architecture and only differ in the temporal stride. Details of the network settings are described in Sec.~\ref{detail}.

Based on $f(v_k)$ and $f'(v)$, we refer to the augmentation parameters in $\tau_p^k$ to calculate the dense spatio-temporal position correspondence. Specifically, we use $S_k\in\mathbb{R}^{N_c\times N_v}$ to indicate the correspondence result, where $N_c=T_cHW$, $N_v=T_vHW$. $S_k(i,j)$ reveals the correspondence score between $i$-th spatio-temporal grid in $f(v_k)$ and $j$-th grid in $f'(v)$. Essentially, each grid on the feature map is equivalent to a tube covering a certain spatio-temporal area (see Fig. \ref{fig:framework}), and $S_k(i,j)$ is measured by the ratio of the intersection of two tubes over the volume of tube $f(v_k)[i]$: 
\begin{align}
    S_k(i,j)=\frac{\text{inter}(f(v_k)[i],f'(v)[j])}{\text{vol}(f(v_k)[i])},
\end{align}
where $[\cdot]$ denotes grid index, $\text{vol}(\cdot)$ measures the spatio-temporal volume of the given feature tube, and ($\text{inter}(\cdot)$) measures the intersecting volume between two tubes. The detailed computation process is illustrated in the Supplementary Material. In this formulation, the row-wise summation of $S_k$ equals to 1, \emph{i.e.}, $\sum_{j=1}^{N_v}S_k(i,j)=\{1\}^{N_c}$. This indicates that each row in $S_k$ can be treated as a probability distribution that describes the correspondence between $f(v_k)[i]$ and each grid in $f'(v)$.

Therefore, we utilize the calculated correspondence matrix $S_k$ as the reference distribution to guide spatio-temporal region feature contrast for accurate visual content alignment. To be specific, we take $f(v_k)[i]$ as query for illustration. Recall that the InfoNCE loss can be written as the cross-entropy between a prior distribution, \emph{i.e.}, the indicator function, and the feature similarity distribution is given as:
\begin{align}
    \mathcal{L}_{nce}(i) = -\sum_j\mathbb{I}_{ij}\log\frac{\text{sim}(\bm{q}_i,\bm{k}_j)}{\sum_j\text{sim}(\bm{q}_i,\bm{k}_j)},
\end{align}
where $\mathbb{I}_{ij}=1$ if $i=j$ otherwise $\mathbb{I}_{ij}=0$, and $\text{sim}(\cdot,\cdot)=\exp(\cos(\cdot,\cdot)/\tau)$ measures the feature similarity. In our formulation, we replace the prior $\mathbb{I}_{ij}$ with the soft distribution $S_k(i,j)$ for accurate region contrast. Since the correspondence between $v_k$ and clips from other videos naturally equals to 0, we can intuitively enlarge the negative pool by introducing features from other videos. Thus, the spatio-temporal region contrast loss w.r.t. $f(v_k)[i]$ can be formulated as
\begin{align}
    &\mathcal{L}_{rc}(k,i) = -\sum_{j=1}^{N_v}S_k(i,j)\log p_k^{ij},\\
    &p_k^{ij} = \frac{\text{sim}(f(v_k)[i],f'(v)[j])}{\sum_{j=1}^{N_v}\text{sim}(f(v_k)[i],f'(v)[j])+\sum_{\bm{n}}\text{sim}(f(v_k)[i],\bm{n})},
\end{align}
where $\bm{n}$ denotes the negative features sampled from other videos in the mini-batch. In this way, we are able to align the exact corresponding appearance and motion content in videos.

\subsection{Low-level Shortcut Elimination}
\label{sec:shortcut}
However, local-global spatio-temporal correspondence for region feature contrast, can exist in the form of a ``shortcut'' that relies merely on low-level statistics, \emph{e.g.}, color distribution, to identify the overlapping areas. This shortcut could prevent the model from learning meaningful semantic features. To this end, we aim to mitigate the impact of low-level statistics on the extracted representations.

An intuitive way to solve this problem is by utilizing strong augmentations. However, we find that this is not enough in video domain. Unlike images, the temporal continuity between sampled frames could provide extra cues to learn these shortcuts. For example, the continuous change in illumination helps to determine the corresponding segments in local-global view. It is nontrivial to design augmentations to decouple such low-level information from the final representations. Motivated by adversarial learning, a promising approach is to learn a low-level information estimator from semantically inconsistent samples that share similar low-level statistics. Then, we let the encoder minimize this estimated information.

We note that the color and blur augmentation $\tau_l$ is effective against distortions on low-level statistics. In other words, similar augmentations could generate samples that share similar low-level characteristics. Hence, we define several different intensity levels of $\tau_l$ by constraining the augmentation parameters to a certain range. As such, we could generate frame sequences that possess distinct semantics but similar low-level statistics using the controlled $\tau_l$. Then, we build a mutual information estimator on top of the extracted feature representation for low-level information extraction. Note that there are several ways to approximate the mutual information -- we compare different estimation methods in Sec.~\ref{ablation}. For illustration, we take MINE~\cite{belghazi2018mutual} as example. Following \cite{belghazi2018mutual}, we approximate the mutual information between two variables by
\begin{align}
    I_\Theta(X;Y) = \sup_{\theta\in\Theta}\mathbb{E}_{\mathbb{P}_{XY}}[G_\theta]-\log(\mathbb{E}_{\mathbb{P}_X\otimes\mathbb{P}_Y}[e^{G_\theta}]),
\end{align}
where $X$ and $Y$ is the feature representations extracted by encoder $f$, $G_\theta:\mathcal{X}\times\mathcal{Y}\rightarrow\mathbb{R}$, which is parameterized by a neural network with $\theta\in\Theta$. We instantiate $G_\theta$ as a two-layer MLP. We regard the features of sample pairs generated from the same intensity-level of $\tau_l$ as the joint distribution $\mathbb{P}_{XY}$, and features of arbitrary sample pairs as the marginal $\mathbb{P}_X\otimes\mathbb{P}_Y$. During training, we formulate the learning objective as:
\begin{align}
    \mathcal{L}_{mi} = \min_f\max_\theta \mathbb{E}_{\mathbb{P}_{XY}}[G_\theta]-\log(\mathbb{E}_{\mathbb{P}_X\otimes\mathbb{P}_Y}[e^{G_\theta}]).
    \label{eq:mi}
\end{align}
We maximize Eq.~\ref{eq:mi} in regards to the MLP parameters $\theta$ to obtain a reliable low-level information extractor, but reverse the gradient back-propagated to the encoder $f$ to minimize Eq.~\ref{eq:mi}. With the learned low-level information estimator $G_\theta$, we further apply it to the aforementioned local-global pairs, $f(v_k)$ and $f'(v)$, to minimize the low-level shortcut by optimizing $f$, but not update $\theta$. In this way, we minimize the impact of low-level statistics on the spatio-temporal region feature contrast, and facilitate detailed semantic alignment.

\subsection{Local-global Temporal Dependency}

Now, we have learned robust clip features from the detailed region semantic contrast, the remaining task is to bridge the gap between clip-level and video-level representations. Considering that there exist the internal temporal relationships between the sampled local clips (which are naturally contained in the global video), we propose to model the temporal order dependency between $f(v_k),k=\{1,2,...,K\}$ and $f'(v)$ to enhance video-level understanding.

Similar to Sec. \ref{sec:shortcut}, we also use mutual information to measure the local-global temporal order dependency. The target is to maximize the mutual information between correctly ordered clip-level features and the video-level representation. Mathematically, we denote the sequentially ordered clip features as $\overline{f}(v)=\left[f(v_1)\circ f(v_2)\circ\cdots\circ f(v_K)\right]$, and the arbitrarily ordered features as $\widetilde{f}(v)$. To model the temporal dependency, we regard $\overline{f}(v)$ and $f'(v)$ as sampled from the joint distribution $\mathbb{P}_{XY}$, and $\widetilde{f}(v)$ and $f'(v)$ as sampled from the marginal distribution $\mathbb{P}_X\otimes\mathbb{P}_Y$. In this formulation, the learning objective can be written as
\begin{align}
    \mathcal{L}_{td} = \max_{f,\psi}\mathbb{E}_{\mathbb{P}_{XY}}[G_\psi]-\log(\mathbb{E}_{\mathbb{P}_X\otimes\mathbb{P}_Y}[e^{G_\psi}]),
\end{align}
where $G_\psi$ is the mutual information estimation head. There exist several alternatives to instantiate $G_\psi$, and we discuss this in Sec.~\ref{sec:ablation}.

It is worth noting that there are some previous works using temporal order to build pretext tasks for self-supervised learning~\cite{misra2016shuffle,lee2017unsupervised,xu2019self}. The major difference is that our approach incorporates the video-level feature to determine whether the clips are correctly ordered, while \cite{misra2016shuffle,lee2017unsupervised,xu2019self} have no access to the global feature. In this way, our formulation could avoid the ambiguity problem when encountering the temporal structure that cannot be determined solely by local clips. For example, in a complex gymnastic scene, it is difficult to determine the temporal order of gymnastic actions only with local clips. But with reference to the global video feature, it is practical to reach the correct order. Thus, our local-global mutual temporal order constraint is be a better way to embed the video-level temporal structures into extracted representations.

\section{Experiment}

\subsection{Datasets}

We use 4 video action recognition datasets, Kinetics-400~\cite{carreira2017quo}, UCF-101~\cite{soomro2012ucf101},  HMDB-51~\cite{kuehne2011hmdb} and Diving-48~\cite{li2018resound}. \textbf{Kinetics-400}~\cite{carreira2017quo} is a large-scale dataset consisting of 240K video clips with 400 human action classes. \textbf{UCF-101}~\cite{soomro2012ucf101} contains over 13k clips covering 101 action classes. \textbf{HMDB-51}~\cite{kuehne2011hmdb} covers 51 action categories and around 7K annotated clips. \textbf{Diving-48}~\cite{li2018resound} contains 48 different diving actions, which mainly vary in motion patterns and share almost similar backgrounds. In our experiments, we use the training set of UCF-101 or Kinetics-400 for self-supervised pretraining. For the downstream tasks, following~\cite{benaim2020speednet,han2020memory,dave2021tclr}, we use split 1 of UCF-101 and HMDB-51, and V1 test set of Diving-48 for evaluation.

\subsection{Implementation Details}
\label{detail}
\noindent\textbf{Self-supervised Pretraining.} For global video input, we sparsely sample 16 frames with weak spatial cropping. For local clips input, we constrain the temporal cropping parameters to make $K$ 16-frame clips uniformly distributed (approximately) in the video. The local clips are spatially cropped within the global view to ensure position-wise correspondence. For low-level augmentations, we define a set of color jitter and Gaussian blur parameters to form different intensity-level transformations. We resize the input frame sequence into $16\times 112\times 112$, and use R3D-18~\cite{hara2017learning} as the video encoder. For local clip feature extraction, we follow the default setting and the feature resolution is $2\times 4\times 4$. For global video feature extraction, we set the temporal stride of the last 3 stages to 1, so that the feature resolution is $8\times 4\times 4$. We calculate the spatio-temporal correspondence matrix between local and global features maps based on the cropping and flipping parameters for optimization. 

In terms of training settings, we use batchsize 128, and set the number of local clips $K$ to 4 by default. We train our model on UCF-101 for 200 epochs, and on Kinetics-400 for 100 epochs. We use Adam optimizer with initial learning rate of $10^{-3}$, weight decay $10^{-5}$. The learning rate is decayed by 10 at 70 epochs for Kinetics-400, and 150 epochs for UCF-101.

\noindent\textbf{Action Recognition.} We load the pretrained video encoder parameters except the last fully-connected layer. There are two protocols: 1) End-to-end \textit{finetune} the whole network with action labels; 2) Freeze the encoder, only train the linear classifier, also known as \textit{linear probe}. For evaluation, we follow~\cite{xu2019self,wang2020self} to uniformly sample 10 clips for each video, which are center cropped and resized to $112\times 112$. We average the softmax probability of each clip as final prediction, and report the Top-1 accuracy.

\noindent\textbf{Video Retrieval.} We directly use the pretrained model to extract video features without finetuning. Following~\cite{xu2019self,luo2020video}, we regard videos in test set as query, and retrieve nearest neighbors from training set. Similar to action recognition, we average the feature of ten uniformly sampled clips as the global representation. We report Top-k recall R@k.

\subsection{Comparison with Existing Works}

\noindent\textbf{Action Recognition.}
We first present the comparison between our method with recent video representation learning approaches on action recognition in Table~\ref{tab:recognition}. We report Top-1 accuracy on UCF-101 and HMDB-51 under \textit{linear probe} and \textit{finetune}. We exclude the methods that use different evaluation settings and much deeper backbone like~\cite{qian2020spatiotemporal,li2021motion,feichtenhofer2021large}, or those that rely on audio and text modalities like~\cite{asano2020labelling,patrick2020multi}. In Table~\ref{tab:recognition}, we use `V+F' to denote the use of both RGB and optical flow in the self-supervised pretraining stage. All evaluation results are obtained using only RGB at test time.

\begin{table}[t]
\centering
\small
\scalebox{0.9}{
    \begin{tabular}{c|cccccc|cc}
        \hline
        Method & Backbone & Pretrain Dataset & Frames & Res. & Mod. & Freeze & UCF-101 & HMDB-51 \\\hline
        CBT~\cite{sun2019learning} & S3D & Kinetics-600 & 16 & 112 & V & \Checkmark & 54.0 & 29.5 \\
        MemDPC~\cite{han2020memory} & R3D-34 & Kinetics-400 & 40 & 224 & V & \Checkmark & 54.1 & 30.5\\
        RSPNet~\cite{chen2021rspnet} & R3D-18 & Kinetics-400 & 16 & 112  & V & \Checkmark & 61.8 & 42.8 \\
        MLRep~\cite{qian2021enhancing} & R3D-18 & Kinetics-400 & 16 & 112 & V & \Checkmark & 63.2 & 33.4 \\
        %Time-Eq~\cite{jenni2021time} & R3D-18 & UCF-101 & 16 & 128 & V & \Checkmark & 74.1 & 47.5 \\
        CoCLR~\cite{han2020self} & S3D & Kinetics-400 & 32 & 128 & V+F & \Checkmark & \textbf{74.5} & \textbf{46.1} \\
        \hdashline
        Ours & R3D-18 &  Kinetics-400 & 16 & 112 & V & \Checkmark & 71.2 & 44.1 \\
        \hline
        \hline
        VCP~\cite{luo2020video} & R3D & UCF-101 & 16 & 112  & V & \XSolidBrush & 66.3 & 32.2 \\
        % PRP~\cite{yao2020video} & R(2+1)D & UCF-101 & 16 & 112 & V & \XSolidBrush & 72.1 & 35.0\\
        TCLR~\cite{dave2021tclr} & R3D-18 & UCF-101 & 16 & 112 & V & \XSolidBrush & 82.4 & 52.9 \\
        LSFD~\cite{behrmann2021long} & R3D & UCF-101 & 32 & 112 & V & \XSolidBrush & 77.2 & 53.7 \\
        %Time-Eq~\cite{jenni2021time} & R3D-18 & UCF-101 & 16 & 128 & V & \XSolidBrush & 83.7 & 62.1 \\
        STS~\cite{wang2020statistic} & R(2+1)D & UCF-101 & 16 & 112 & V+F & \XSolidBrush & 77.8 & 40.7 \\
        CoCLR~\cite{han2020self} & S3D & UCF-101 & 32 & 128 & V+F & \XSolidBrush & 81.4 & 52.1 \\
        \hdashline
        Ours & R3D-18 & UCF-101 & 16 & 112 & V & \XSolidBrush & \textbf{83.3} & \textbf{53.2} \\
        \hline
        Pace~\cite{wang2020self} & R(2+1)D & Kinetics-400 & 16 & 112  & V & \XSolidBrush & 77.1 & 36.6\\
        MemDPC~\cite{han2020memory}& R3D-34 & Kinetics-400 & 40 & 224 & V & \XSolidBrush & 78.1 & 41.2\\
        VideoMoCo~\cite{pan2021videomoco} & R(2+1)D & Kinetics-400 & 32 & 112 & V & \XSolidBrush & 78.7 & 49.2\\
        RSPNet~\cite{chen2021rspnet} & R(2+1)D & Kinetics-400 & 16 & 112  & V & \XSolidBrush & 81.1 &44.6 \\
        TempTrans~\cite{jenni2020video} & R3D-18 & Kinetics-400 & 16 & 112  & V & \XSolidBrush & 79.3 & 49.8 \\
        MLRep~\cite{qian2021enhancing} & R3D-18 & Kinetics-400 & 16 & 112 & V & \XSolidBrush & 79.1 & 47.6 \\
        ASCNet~\cite{huang2021ascnet} & R3D-18 & Kinetics-400 & 16 & 112 & V & \XSolidBrush & 80.5 & 52.3 \\
        TCLR~\cite{dave2021tclr} & R3D-18 & Kinetics-400 & 16 & 112 & V & \XSolidBrush & 84.1 & 53.6 \\
        Time-Eq~\cite{jenni2021time} & R3D-18 & Kinetics-400 & 16 & 128 & V & \XSolidBrush & 87.1 & \textbf{63.6} \\
        STS~\cite{wang2020statistic} & S3D-G & Kinetics-400 & 64 & 224 & V+F & \XSolidBrush & \textbf{89.0} & 62.0 \\
        CoCLR~\cite{han2020self} & S3D & Kinetics-400 & 32 & 128 & V+F & \XSolidBrush & 87.9 & 54.6 \\
        \hdashline
        Ours & R3D-18 & Kinetics-400 & 16 & 112 & V & \XSolidBrush & 85.5 & 55.1 \\
        Ours* & R3D-18 & Kinetics-400 & 16 & 112 & V & \XSolidBrush & 88.1 & 56.4 \\
        Ours & R3D-18 & Kinetics-400 & 16 & 224 & V & \XSolidBrush & 88.3 & 59.3 \\
        Ours & R3D-34 & Kinetics-400 & 16 & 112 & V & \XSolidBrush & 87.1 & 58.4 \\
        \hline
    \end{tabular} 
    }    
    \caption{Comparison results for action recognition downstream task. We provide the training setting of each method, including backbone encoder, pretraining dataset, spatio-temporal resolution and the modality, where `V' means RGB frames, `F' means optical flow. We use freeze (tick) to indicate linear probe, while no freeze (cross) denotes end-to-end fine-tuning. For fairness, note that we exclude methods that use different evaluation settings, much deeper backbones or other modalities like audio and text. And `*' denotes 200 epochs pretraining on Kinetics-400.}
    \vspace{-8mm}
    \label{tab:recognition}
\end{table}
Under \textit{linear probe}, our method outperforms other RGB-only approaches by a large margin. The superiority over RSPNet~\cite{chen2021rspnet}, which integrates temporal pretext task with contrastive learning, demonstrates the effectiveness of our general temporal structure learning scheme. And note that our method also dramatically narrows the gap between RGB-only and RGB-flow based method. This indicates that our method significantly improves the motion pattern modeling. Under \textit{finetune}, our method achieves the best result when pretrained on UCF-101, even surpassing RGB-flow based methods. And when pretrained on Kinetics-400, ours is also comparable with state-of-the-art RGB-flow approaches. Besides, due to limited computation resource, we do not compare with works using very large backbones like~\cite{qian2020spatiotemporal,feichtenhofer2021large}, but we show the ablation in the bottom three lines. The results indicate that our method has potential to scale to longer training epochs, deeper backbone or larger resolution.

\begin{wraptable}{r}{0.5\textwidth}
    \centering
    \begin{tabular}{c|c|c}
    \hline
      Method & Backbone & Top-1  \\
      \hline
      Random Init. & R3D-18  & 13.4 \\
      \hdashline
      RESOUND~\cite{li2018resound} & C3D & 16.4 \\
      TSN~\cite{wang2016temporal} & BN-Inception & 16.8 \\
      Debiased~\cite{choi2019sdn} & R3D-18 & 20.5 \\
      \hdashline
      SimCLR~\cite{chen2020simple} & R3D-18 & 20.1 \\
      TCLR~\cite{dave2021tclr} & R3D-18 & 22.9 \\
      Ours & R3D-18 & \textbf{25.4} \\
    \hline
    \end{tabular}
    \caption{Action recognition results on Diving-48 dataset. We compare different Top-1 accuracy based on V1 action labels.}
    \label{tab:diving}
\end{wraptable}

Besides, we also provide the results on Diving-48~\cite{li2018resound}, a dataset which mainly relies on dynamic motions to distinguish different action categories. We show the comparison results between both supervised (between dashed lines) and self-supervised methods (bottom three) in Table~\ref{tab:diving}. Since the appearance is similar across different videos, the Top-1 accuracy can well reflect the ability in motion understanding. We observe that in this case, semantic label supervision is not that effective, and our method improves the performance by a notable margin. This shows that our learning approach is superior in capturing motion patterns, with less reliance on background information. %much more efficient than the semantic labels. [cannot consider efficiency since computation is not measured / SS normally not that efficient due to pretraining step]

\begin{table}[t]
    \centering
    \begin{tabular}{c|c|cccc|cccc}
        \hline
        \multirow{2}{*}{Method} & \multirow{2}{*}{Backbone} & \multicolumn{4}{c|}{UCF-101} & \multicolumn{4}{c}{HMDB-51} \\
        \cline{3-10}
         & & R@1 & R@5 & R@10 & R@20 & R@1 & R@5 & R@10 & R@20 \\
        \hline
        PRP~\cite{yao2020video} & R3D & 22.8 & 38.5 & 46.7 & 55.2 & 8.2 & 25.8 & 38.5 & 53.3 \\
        Pace~\cite{wang2020self} & R3D-18 & 23.8 & 38.1 & 46.4 & 56.6 & 9.6 & 26.9 & 41.1 & 56.1 \\
        MemDPC~\cite{han2020memory} & R3D & 20.2 & 40.4 & 52.4 & 64.7 & 7.7 & 25.7 & 40.6 & 57.7 \\
        MLRep~\cite{qian2021enhancing} & R3D-18 & 39.6 & 57.6 & 69.2 & 78.0 & 18.8 & 39.2 & 51.0 & 63.7 \\
        PCL~\cite{tao2020selfsupervised} & R3D-18 & 40.5 & 59.4 & 68.9 & 77.4 & 16.8 & 38.4 & 53.4 & \textbf{68.9} \\
        STS~\cite{wang2020statistic} & R3D-18 & 38.3 & 59.9 & 68.9 & 77.2 & 18.0 & 37.2 & 50.7 & 64.8 \\
        CoCLR~\cite{han2020self} & S3D & \textbf{53.3} & 69.4 & \textbf{76.6} & \textbf{82.0} & \textbf{23.3} & 43.2 & 53.5 & 65.5 \\
        \hdashline
        Ours & R3D-18 & 52.7 & \textbf{70.3} & 76.2 & 81.1 & 22.9 & \textbf{45.8} & \textbf{56.1} & 64.2 \\
        \hline
    \end{tabular}
    \caption{Comparison results for video retrieval downstream task. We report R@k (k=1,5,10,20) on UCF-101 and HMDB-51 datasets.}
    \vspace{-4mm}
    \label{tab:retrieval}
\end{table}

\noindent\textbf{Video Retrieval.}
Table~\ref{tab:retrieval} shows the comparison on video retrieval with R@k. The model is pretrained on UCF-101. Our method remarkably outperforms most RGB-based approaches. Note that some methods, especially PCL~\cite{tao2020selfsupervised}, achieve impressive results when $k$ increases to 20. This is because when $k$ is large, it becomes likely to rely on background as a shortcut to retrieve videos of the same category. Though STS~\cite{wang2020statistic} and CoCLR~\cite{han2020self} adopt RGB and optical flow, we reach comparable or even better performance. This again demonstrates that our integration of detailed local feature alignment and general long-term temporal modeling is effective in enhancing motion pattern modeling without resorting to motion biased input data.

\noindent\textbf{Visualization Analysis.}
\begin{figure}[t]
    \centering
    \subfigure[Results of baseline video contrastive learning.]{
    \includegraphics[width=0.15\linewidth]{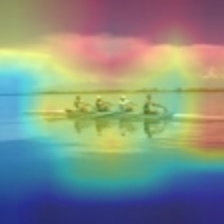}
    \includegraphics[width=0.15\linewidth]{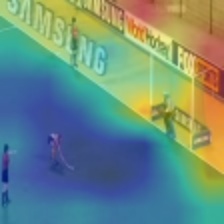}
    \includegraphics[width=0.15\linewidth]{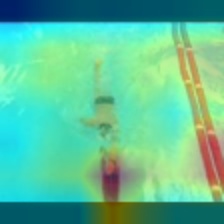}
    \includegraphics[width=0.15\linewidth]{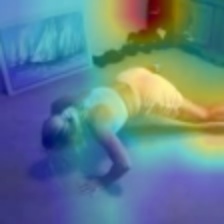}
    \includegraphics[width=0.15\linewidth]{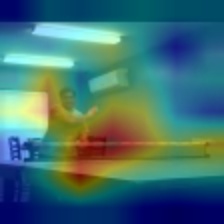}
    \includegraphics[width=0.15\linewidth]{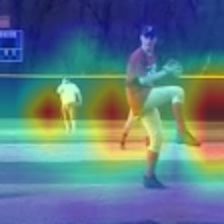}
    }
    \subfigure[Results of our learning approach.]{
    \includegraphics[width=0.15\linewidth]{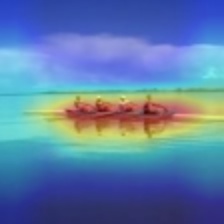}
    \includegraphics[width=0.15\linewidth]{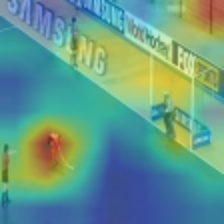}
    \includegraphics[width=0.15\linewidth]{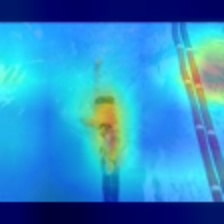}
    \includegraphics[width=0.15\linewidth]{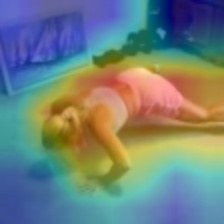}
    \includegraphics[width=0.15\linewidth]{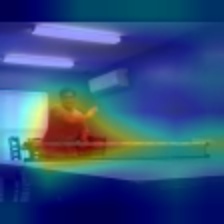}
    \includegraphics[width=0.15\linewidth]{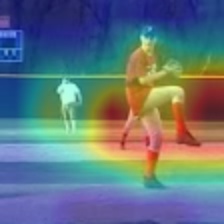}
    }
    \caption{CAAM visualization of spatio-temporal feature maps. We compare the results of our method and contrastive learning baseline. Ours focuses on the moving objects while the baseline inclines to emphasize background regions.}
    \label{fig:vis}
    \vspace{-3mm}
\end{figure}
We also show some visualization results to analyze the learned feature representations in Fig.~\ref{fig:vis}. We employ class-agnostic activation maps (CAAM)~\cite{baek2020psynet} to reveal the spatio-temporal distributions of the extracted features. Generally, the vanilla contrastive learning based on SimCLR~\cite{chen2020simple} leads the model to focus on some representative background cues, \emph{e.g.}, the soccer field, swimming pool and fitness equipment. On the contrary, our pretrained model focuses on the moving foregrounds that contain actions, like the moving human body and moving boat.

\subsection{Ablation Study}

In this section, we provide several ablation studies to analyze our video representation learning framework. If not specifically mentioned, all models are pretrained on UCF-101 for 150 epochs, with R3D-18 as backbone.

\label{sec:ablation}
\begin{table}
    \centering
    \begin{tabular}{ccc|cc}
    \hline
      $\text{ }K\text{ }$ & $\text{ }T_v\text{ }$ & $\text{ }T_v/KT_c\text{ }$ &  UCF-101  &  HMDB-51  \\
      \hline
      1 & 2 & 1.0 & 47.8 & 26.3 \\
      \hdashline
      2 & 2 & 0.5 & 50.1 & 27.5 \\
      2 & 4 & 1.0 & 53.5 & 28.9 \\
      3 & 2 & 0.33 & 51.5 & 28.1 \\
      3 & 4 & 0.67 & 57.7 & 29.7 \\
      3 & 8 & 1.33 & 58.3 & 29.9 \\
      4 & 4 & 0.5 & 54.2 & 28.4 \\
      4 & 8 & 1.0 & \textbf{60.1} & \textbf{31.3} \\
    \hline
    \end{tabular}
    \caption{Ablation study on local-global sampling. We show the results with different clip numbers and the temporal resolution of global video feature. The first line equals to baseline. We report linear probe Top-1 accuracy on UCF-101 and HMDB-51.}
    \vspace{-4mm}
    \label{tab:local-global}
\end{table}
\noindent\textbf{Local-global Sampling.}
We first explore the impact of local-global settings. Two aspects were investigated, one is the number of local clips $K$, the other is the global video feature temporal resolution $T_v$, which is obtained by adjusting temporal convolution stride. We show the results in Table~\ref{tab:local-global}. By varying the number of local clips $K$ from 1 to 4, we find that having more local clips tend to improve the performance due to more fine-grained feature alignment. And it is worth noting that when the ratio $T_v/KT_c<1$, the granularity of local-global correspondence becomes too coarse, which constricts the performance. Overall, accurate spatio-temporal region correspondence does provide reliable reference for appearance and motion pattern matching, and significantly improves action recognition.

\begin{table}
    \centering
    \begin{tabular}{ccc|cc}
    \hline
      \#B/C/S & \#H & \#G & UCF-101 &  HMDB-51  \\
      \hline
      2 & 2 & 2 & 55.2 & 28.4 \\
      4 & 2 & 4 & \textbf{60.1} & \textbf{31.3} \\
      4 & 4 & 4 & 58.3 & 29.5 \\
      5 & 2 & 5 & 59.4 & 30.7 \\
    \hline
    \end{tabular}
    \caption{Ablation study on low-level augmentation settings. \# denotes the number of intensity levels in Brightness, Contrast, Saturation, Hue and Gaussian Blur. We report linear probe Top-1 accuracy on UCF-101 and HMDB-51.}
    \vspace{-5mm}
    \label{tab:low}
\end{table}
\noindent\textbf{Low-level Augmentation Levels.}
We also explore the setting of the intensity levels on low-level augmentations. We follow conventional implementations: For color jitter, there are controllable parameters of brightness, contrast, saturation and hue, which are set as (B,C,S,H)=(0.4,0.4,0.4,0.1) by default~\cite{han2020self,pan2021videomoco}. For Gaussian blur, we control the radius and sigma. We set different intensity levels for each controllable parameter as Table~\ref{tab:low}. Note that since B,C,S are set to the same as default, we also set the same number of levels for them. The total number of predefined intensity levels equals to the number of permutations across all parameters, 
%product of each , 
\emph{i.e.}, 32 for the first row, 512 for the second row, etc. For consistency, in each iteration, we randomly sample 32 intensity levels from all possible levels, resulting in 32 groups of features that share similar low-level statistics for mutual information minimization. We observe that too few or too many levels both leads to performance drop. This is because more levels leads to less difference between different groups, while fewer levels means more difference within each group. We conclude that there exists a trade-off that requires balancing to achieve the best possible training.

\begin{table}
\begin{minipage}{0.45\linewidth}
    % \centering
    % \begin{tabular}{c|cc}
    % \hline
    %   Estimation  &  mini-Kinetics  &  UCF-101  \\
    %   \hline
    %   None   & 53.4 & 52.6 \\
    %   \hdashline
    %   MINE~\cite{belghazi2018mutual} & 64.1 & 58.8 \\
    %   JS~\cite{nowozin2016f}  & 64.5 & 57.6 \\
    %   InfoNCE~\cite{oord2018representation}  & \textbf{67.5} & \textbf{60.1} \\
    %   CLUB~\cite{cheng2020club}  & 61.4 & 56.4 \\
    % \hline
    % \end{tabular}
    % \caption{Ablation study on low-level mutual information estimation. None means the baseline without mutual information minimization.}
    % \label{tab:variants}
    \centering
    \begin{tabular}{c|cc}
    \hline
      Head  &  UCF-101  &  HMDB-51  \\
      \hline
      None   & 54.5 & 28.3 \\
      VCOP~\cite{xu2019self} & 57.6 & 29.4 \\
      \hdashline
      MLP   & 58.8 & 30.1 \\
      GRU  & 60.1 & \textbf{31.3} \\
      GRU+MLP  & \textbf{60.5} & 31.0 \\
    \hline
    \end{tabular}
    \caption{Ablation study on temporal dependency head. None denotes the baseline without temporal constraint, and VCOP follows~\cite{xu2019self} for comparison.}
    \vspace{-5mm}
    \label{tab:tem}
\end{minipage}
\hspace{0.03\linewidth}
\begin{minipage}{0.5\linewidth}
    \centering
    \begin{tabular}{cccc|cc}
    \hline
        $\mathcal{L}_{nce}$ & $\mathcal{L}_{rc}$ & $\mathcal{L}_{mi}$ & $\mathcal{L}_{td}$ & UCF-101 & HMDB-51 \\
        \hline
        \checkmark & & & & 57.1 & 30.3 \\
          & \checkmark & & & 61.4 & 32.5 \\
         & \checkmark & \checkmark & & 66.2 & 38.1 \\
         & \checkmark & & \checkmark & 64.1 & 36.4 \\
        \checkmark & & \checkmark & \checkmark & 62.3 & 35.9 \\
         & \checkmark & \checkmark & \checkmark & 71.2 & 44.1 \\
        \hline
    \end{tabular}
    \footnotesize \caption{Ablation study on all learning objectives. Note that $\mathcal{L}_{nce}$ is the standard contrastive loss function in previous works.}
    \vspace{-5mm}
    \label{ablation}
\end{minipage}
\end{table}

\noindent\textbf{Temporal Dependency Head.}
To further examine the feasibility of temporal dependency head implementation, we compare three typical examples: 1) MLP: concatenate $f'(v)$ and $\overline{f}(v)$ or $\widetilde{f}(v)$ and pass through a MLP to obtain a scalar value. 2) GRU: use GRU to process clip feature sequence, and calculate the cosine similarity between $f'(v)$ and GRU output. 3) GRU+MLP: use GRU to process clip feature sequence, then concatenate with $f'(v)$ and pass through a MLP to get a scalar value. The results are listed in Table~\ref{tab:tem}. Compared with no temporal constraint, all three implementations showed significant improvements. We note that our MLP implementation is similar to VCOP~\cite{xu2019self}, but different in the learning objective. This improvement reveals that introducing the global video feature as reference could enhance temporal structure modeling.

\noindent\textbf{Overall Learning Objectives.}
We finally show the ablation on designed learning objectives in Table~\ref{ablation}, where $\mathcal{L}_{nce}$ is the standard contrastive loss used in existing works. We observe that the integration of $\mathcal{L}_{rc}$ and $\mathcal{L}_{mi}$ significantly outperforms $\mathcal{L}_{ce}$, which indicates that the detailed region contrast with low-level shortcut elimination is more efficient than naive global contrast. Besides, $\mathcal{L}_{td}$, further enables the model to go beyond local clips and establish long-term relationships. The improvement demonstrates that our method well integrates detailed region-level contrast and general long-term temporal perception.

\section{Conclusion}

In this paper, we propose a framework that leverages local clips and global video to enhance self-supervised video representation learning. We employ a set of controllable augmentations to crop local clips and generate groups of samples that share similar low-level attributes. Thereby, we use the soft codes computed from the crop and flip parameters to guide detailed spatio-temporal region contrastive learning, and minimize the mutual information within the same low-level group to avoid shortcuts. Meanwhile, we also incorporate local-global temporal dependency to embed general temporal structures to the extracted video representations. Experiments on downstream tasks of action recognition and video retrieval demonstrate the superiority of our formulation, especially in modeling dynamic motion patterns.

\clearpage
% ---- Bibliography ----
%
% BibTeX users should specify bibliography style 'splncs04'.
% References will then be sorted and formatted in the correct style.
%
\bibliographystyle{splncs04}
\bibliography{egbib}
\end{document}